\documentclass[10pt]{spie}

\usepackage[utf8]{inputenc}
\usepackage{float}
\usepackage{amsmath,amsfonts,amssymb}
\usepackage{multirow}
\usepackage{booktabs}
\usepackage{graphicx}
\usepackage[colorlinks=true, allcolors=blue]{hyperref}

\renewcommand{\supit}[1]{\textsuperscript{\textit{#1}}}

\title{Hybrid Quantum-Classical GANs for the Generation of Adversarial Network Flows}

\author{
Prateek Paudel\supit{a},
Nitin Jha\supit{a},
Abhishek Parakh\supit{a},
Mahadevan Subramaniam\supit{b}\\
\supit{a}Kennesaw State University, Marietta, GA, USA\\
\supit{b}University of Nebraska Omaha, Omaha, NE, USA
}

\authorinfo{Further author information: Send correspondence to Prateek Paudel at 
\href{mailto:ppaudel@students.kennesaw.edu}{ppaudel@students.kennesaw.edu}}

\begin{document}

\maketitle

\begin{abstract}
Classical generative adversarial networks (GANs) have been applied to generate adversarial network traffic capable of attacking intrusion detection systems, but they suffer from shortcomings such as the need for large amounts of high-dimensional datasets, mode collapse, and high computational overhead. In this work, we propose a hybrid quantum-classical GAN (QC-GAN) framework where a variational quantum generator is used to generate synthetic network traffic flows mimicking malicious traffic using latent representations. Instead of sampling classical noise vectors, we encode the latent vector (the hidden features) as a quantum state, which is the basis for claiming more expressive latent representations and reducing computational overhead. A classical discriminator will be trained on real-world datasets (UNSW-NB15) and the proposed QC-GAN-generated fake network flows. In this configuration, the generator aims to minimize the discriminator's ability to distinguish real from fake traffic, while the discriminator aims to maximize its classification accuracy, in an iterative manner. In our attack model, we assume that the attacker is a state actor with access to limited quantum computing power, whereas the discriminator is chosen to be classical, as will likely be the case for most end users and organizations. We test the generated flows using classical intrusion detection system (IDS) models, such as a random forest classifier and a convolutional neural network-based classifier, for their ability to bypass the detection process. This work aims to highlight the possibilities of quantum machine learning as a means of generating advanced attack flows and stress testing classical IDS. Lastly, we further evaluate how hardware-based noise affects these attacks to offer a new perspective on IDS, highlighting the need for a quantum resilient defense system.  
\end{abstract}

\keywords{Quantum Machine Learning(QML),Generative Adversarial Networks(GANs), Intrusion Detection System,Network Traffic Generation}

\section{INTRODUCTION}
\label{sec:intro}  % \label{} allows reference to this section

The increasing complexity of modern cyber-physical systems has increased the importance of reliable intrusion detection systems (IDSs)  to detect sophisticated attacks in computer networks~\cite{11143115}.  Traditionally, many systems have relied on signature based detection methods, which are  effective against known threats but they fail to detect novel attacks. To address this gap, anomaly-based intrusion detection systems have become the main focus of research. These systems use a machine learning (deep learning) based approach to establish a baseline of normal network behaviour, flagging significant deviations as potential attacks~\cite{VIBHUTE20242227}.  

Zolbayar et al. proposed a Generative Adversarial  Network(GAN)-based network intrusion detection system designed to generate adversarial network traffic flows intending to evade Machine learning based Network Intrusion Detection Systems~\cite{zolbayar2022generatingpracticaladversarialnetwork}. Traditional machine learning approaches typically require a substantial amount of high-quality samples and sufficient computing power, making it challenging to adapt to dynamic environments~\cite{Muneer2024}. Despite the enhancements introduced by machine-learning approaches to IDS technology, it still struggles with a growing amount of traffic, its volume and dimensionality, and the compute intensity involved during the training processes. These obstacles spark interest in investigating other computing paradigms with higher computing efficiency and better adaptability~\cite{vieloszynski2024latentqganhybridqganclassical}. This trend is additionally substantiated by quantum-enhanced network proposals, where classical networking protocols such as HTTP are amended to transmit interlaced quantum payloads for efficient, privacy-preserving resource allocation ~\cite{10821224}.

Quantum computing is one such option available, providing functionality that is fundamentally different from classical computing. Using the principles of superposition and entanglement, quantum computing allows multiple processes to be performed simultaneously using a single bit, which is not possible using classical computers~\cite{Schuld2014IntroQML, Mirza2014CGAN, Huang2021QGAN, article1,math12213318}.
The worldwide scalability of this paradigm would probably require a multi-tier network topology that combines ground stations with airborne and satellite-based links in order to circumvent physical distance constraints ~\cite{Jha_2025}.

Quantum generators via Variational Quantum Circuits (VQCs) have already demonstrated their ability to provide a more expressive latent space than classical generators. The standard GAN consists of multiple samples that are generated from a noise distribution, such as a Gaussian, which requires a strong neural network to map this latent space to generate complex geometric figures. A VQC, on the other hand, functions within a Hilbert space that can grow exponentially with the increase in its qubits, facilitating the representation of more complex probability distributions directly on the amplitudes contained within a quantum state.

This exponential representation capability provides quantum and hybrid quantum-classical generators with the capability to represent more complex correlations than those generated by purely classical architectures, hence prompting their investigation within next-generation GANs. These representational capabilities make it likely that quantum-boosted generators could successfully address some of the issues faced by classical GANs. This provides a very promising route to being able to incorporate VQCs into GAN architectures~\cite{goh2025quantumenhancedgenerativeadversarialnetworks, nguemto2022reqganoptimizedadversarialquantum, rao2023learningharddistributionsquantumenhanced, Shu2024VQCGAN}.  

There has been growing interest in employing Quantum Generative Adversarial Networks (QGANs) in intrusion detection applications due to their capability to capture complex patterns within threat distributions with drastically reduced parameters compared to standard deep architectures. The work by He et al. (2025) highlighted that QGAN-assisted data augmentation can enhance the performance potential of hybrid classifiers combining both quantum and classical computing resources while employing approximately half the required parameters of standard DCGAN architectures, validating that quantum latent spaces can indeed optimize model sizes without hindering detection accuracy~\cite{he2025qganbaseddataaugmentationhybrid}.  
Stein et al. (2021) demonstrated that QC-GAN architectures can outperform classical GANs with fewer parameters and training iterations by exploiting the expressiveness of variational quantum circuits~\cite{Stein_2021}. Capitalizing on such efficiency gains, Hammami, Cherkaoui, and Wang (2025) developed a VQC-controlled QGAN incorporating sequential data injection to accomplish state-of-the-art anomaly detection accuracy on multiple variables within network traffic while employing merely 80 trainable parameters, which remains one of the leanest high-performance IDS architectures to date\cite{Hammami2025QGANAnomaly}.

Finally, to complement this accomplishment within a second publication, Hammami et al. (2025) also developed a Quantum-GRU GAN, where the temporal dependencies within traffic patterns were successfully described by a quantum-boosted recurrent structure, which helped stabilize QGAN-assisted model generation and identified range-wise correlations outside standard GAN architectures~\cite{hammami2025quantumgrugan}. Rahman et al. (2024) continued to experimentally confirm QGAN effectiveness on intrusion detection on the level of packets, concluding that quantum-assisted generators more realistically approximated very high-dimensional and generally very non-linear attack patterns within traffic, promoting more accurate classification results concerning adversarial distorted traffic samples~\cite{rahman2024quantumintrusion}.
Lastly, incorporating error correction schemes directly within quantum communication protocols, for example, by integrating CSS codes with multiple communication stages, reduces NISQ hardware errors while retaining the low computation overhead necessary for real-time IDS ~\cite{Jha2024}.

\section{Methodology}
\label{sec:Methodoloy}

\subsection{Data Preparation and Analysis}
\label{sec:title}
The UNSW-NB15 dataset is used to test quantum-enhanced adversarial traffic generation. The UNSW-NB15 dataset was created by the Australian Centre for Cyber Security (ACCS) to overcome some of the problems with older datasets which did not adequately capture modern network traffic with both benign and different types of attack~\cite{Moustafa2015UNSWNB15,Moustafa2016EvaluationUNSW,Moustafa2017GeometricArea,Moustafa2017BigData,Sarhan2020NetFlow}. The UNSW-NB15 network traffic dataset contains 49 features that define network traffic at different levels, such as flow, packet, and content. These features can be used to detect different types of network attacks, including nine different attack categories, and normal network traffic. Machine learning and anomaly-based intrusion detection systems can use these features to detect different and complex network activities. The UNSW-NB15 network traffic dataset can also be represented using different formats, such as NetFlow. Although using the entire feature set, which is 49-dimensional, is useful for classical machine learning models, it is not possible to use this feature set to directly map it to a quantum generator. This is due to the qubit and quantum circuit constraints of current quantum devices. We use a two-stage feature selection approach to select features from this feature set, which can be used to map to a quantum generator.  

\subsubsection{Stage 1: Coarse Feature Screening}
To select features from the feature set, we use a multi-view ensemble feature selection approach, which combines three different views to measure feature importance. We use Random Forest (RF) Gini impurity to measure non-linear relationships, Mutual Information(MI) to measure statistical dependencies, and L1-regularized logistic regression(L1) to measure feature importance using coefficient selection. We normalize each feature score and use a weighted aggregation to compute the ensemble feature.

$\mathbf{Result} = 0.35 \cdot \text{RF} \;+\; 0.35 \cdot \text{MI} \;+\; 0.30 \cdot \text{L1} \;$

This helps us ensure that we are not choosing multiple features to describe one characteristic of network activity and also helps prevent choosing features that contribute little to the basic structure of the data. Ultimately, we arrive at a final list of four features that capture all the major aspects of network activity: connection timing, protocol state evolution, volume, and packets.

\begin{table}[t]
\centering
\caption{Selected top 8 features from the UNSW-NB15 dataset.}
\renewcommand{\arraystretch}{1.25}
\setlength{\tabcolsep}{8pt}
\begin{tabular}{|p{3.2cm}|p{10.8cm}|}
\hline
\textbf{Feature} & \textbf{Description} \\
\hline
\textbf{id} & A distinct flow identifier that uniquely distinguishes one network flow from another. \\
\textbf{tcprtt} & Measures the TCP round-trip time, including the three-way handshake and the return packet's transit latency. \\
\textbf{synack} & Captures the SYN/ACK time interval during connection establishment, reflecting early-stage handshake behavior. \\
\textbf{ct\_state\_ttl} & A combined indicator of connection state and TTL patterns, useful for detecting abnormal packet lifecycle signatures. \\
\textbf{sbytes} & Represents the total number of bytes sent from the source, helping identify anomalous or excessive outgoing data volumes. \\
\textbf{state} & Denotes the final connection status (e.g., success, reset), which aids in identifying irregular or failed communication attempts. \\
\textbf{sload} & Indicates the source-side load or bandwidth consumption, often revealing bursty or attack-related traffic behavior. \\
\textbf{smean} & The average size of transmitted packets, helping characterize irregular packet-size distributions in malicious flows. \\
\hline
\end{tabular}
\end{table}

\subsubsection{Stage 2: Fine-Grained Selection}
In this second step,  Principal Component Analysis (PCA) is used as a selection method to further perform dimensionality reduction on the original feature set. By examining how each feature is aligned relative to the total variance of the data, PCA is employed to select the features that capture complementary aspects of network behavior instead of redundant information. This prevents the selection of several features that describe the same network behavior and also drops those features that do not significantly contribute to the overall structure of the network behavior. This way, a four-feature subset is obtained that captures the fundamental aspects of network behavior: timing of connections, protocol state evolution, traffic volume, and packet characteristics.

The final quantum feature set is composed of synack, ct\_state\_ttl, sbytes, and smean.
This feature set is both expressive and constrained enough to capture the original variability of the UNSW-NB15 dataset while being compatible with the four-qubit quantum generator employed in our framework. This limits the input dimensionality to be within the range that is both expressive and feasible for the quantum model without the need for additional qubits or circuit depths.

\begin{figure}[H]
    \centering
    \includegraphics[width=0.75\textwidth]{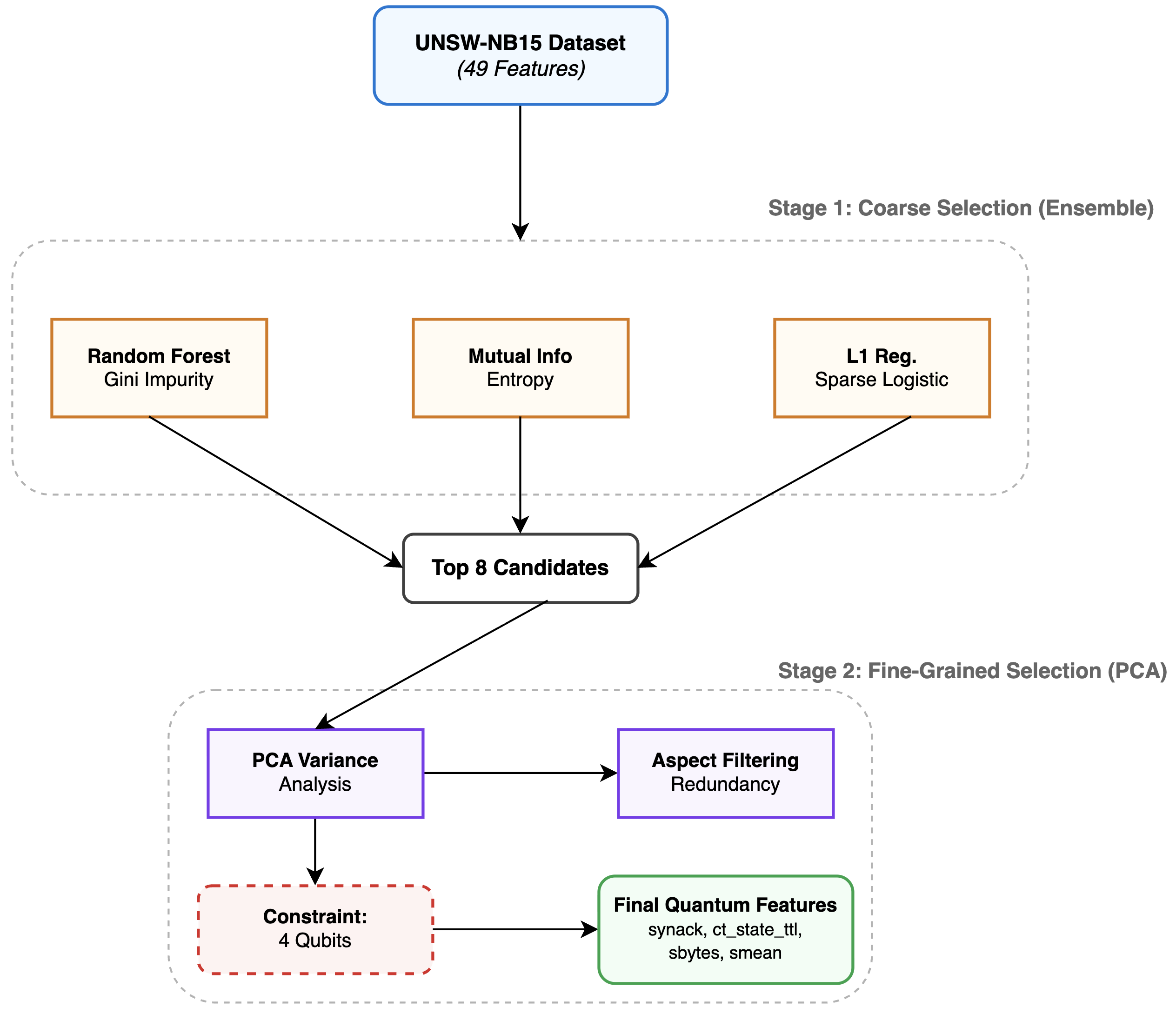}
    \caption{Two-stage feature selection pipeline for quantum encoding with qubit constraints.
    Coarse screening of candidate features is performed by ensemble and sparsity-driven methods in Stage~1.
    Stage~2 applies variance and redundancy analysis to select a final subset consistent with the four-qubit quantum generator.}
    \label{fig:feature_select}
\end{figure}

\subsection{Model Architecture}

The framework proposed combines aspects of quantum and classical computing within a GAN architecture that is similar to that used in real-world applications. The generator uses a variational quantum circuit with a limited number of qubits, representing a limited level of quantum access that an attacker could have. The discriminator, meanwhile, remains classical, similar to how intrusion detection systems are used in real applications.

Contrary to classical GANs, in which the generator is represented using a deep neural network with multiple fully connected layers, the proposed quantum generator works in a high-dimensional Hilbert space determined by the degree of qubits. This helps the generator capture complex correlations in network traffic using far fewer parameters. It takes a latent noise vector drawn from a standard normal distribution, which is the stochastic seed for the generation of network flows.

\begin{figure}[H]
    \centering
    \includegraphics[width=\linewidth]{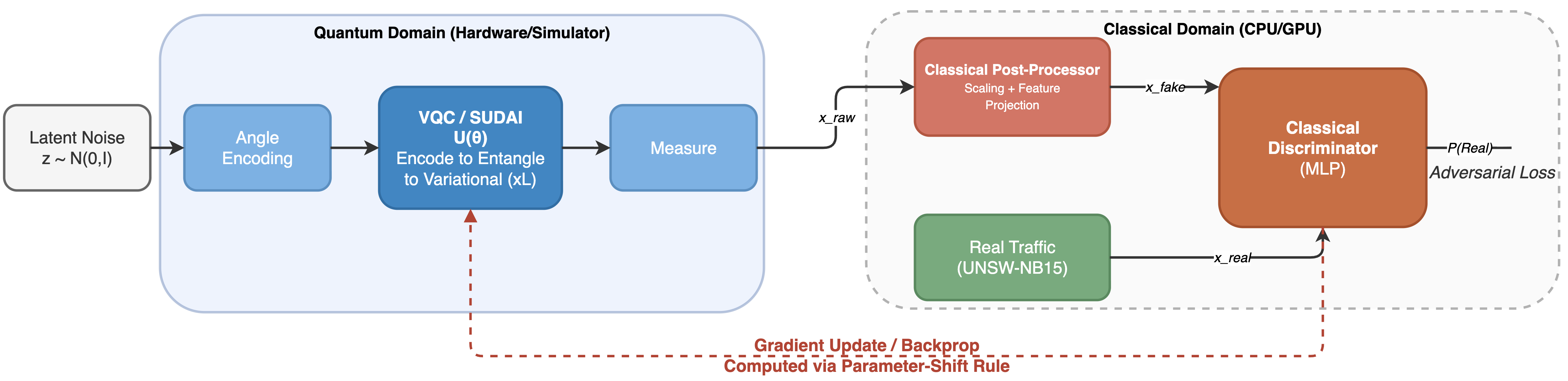}
    \caption{Architecture of the Hybrid Quantum-Classical GAN (QC-GAN). The framework consists of a Quantum Generator utilizing a Variational Quantum Circuit (VQC) with Successive Unitary Data Injection (SUDAI) encoding, a classical post-processing network, and a Classical Discriminator based on a Multi-Layer Perceptron (MLP) to classify real traffic from the UNSW-NB15 dataset and synthetic samples.}
    \label{fig:architecture}
\end{figure}

The latent vector is represented as a quantum state via a series of parametric quantum gates, followed by the unitary transformation of the quantum state. A measurement of the quantum state produces a low-dimensional representation of the classical data, representing synthetic network traffic features. In order to better match real data statistics, additional refinement of the measurement of quantum outputs is optionally applied by a classical post-processing network. A neural network representing the classical discriminator takes either real samples of network traffic data from the UNSW-NB15 dataset or synthetic data from the quantum generator as input. It then produces a scalar value that is a probability of real or generated network traffic data. Both the generator and discriminator are used in an adversarial training process where the generator tries to generate more real attack traffic, and the discriminator tries to maximize its own classification accuracy. Figure~\ref{fig:architecture} illustrates the architecture of our proposed Hybrid Quantum-Classical GAN.

\subsubsection{Encoding Layer}

The function of this layer is to establish a connection between the classical and the quantum   by embedding a continuous latent vector into a quantum generator.

Given a classical latent noise vector sampled from a standard normal distribution, the quantum generator employs angle encoding to transform this vector into quantum states. This is done by maintaining a small circuit depth, which is suitable for the hardware and continuous variables.
Each element of the latent vector is responsible for rotating a single qubit by an angle determined by that element of the latent vector. Thus, the classical random variable is encoded in the quantum amplitudes.

The latent vector can be defined as:
\begin{equation}
\mathbf{z} \in \mathbb{R}^N, \quad \mathbf{z} \sim \mathcal{N}(0, I)
\end{equation}
where $N$ is the total number of qubits used in the quantum generator. Before encoding, each latent component is normalized to the range $[-1,1]$ to provide bounded rotation angles and reliable quantum state preparation on near-term quantum processors.

The continuous latent variables are embedded via single-qubit rotation gates regarding the Y-axis. The unitary operation representing an $R_Y$ rotation is given by:
\begin{equation}
R_Y(\theta) =
\begin{pmatrix}
\cos\left(\frac{\theta}{2}\right) & -\sin\left(\frac{\theta}{2}\right) \\
\sin\left(\frac{\theta}{2}\right) & \cos\left(\frac{\theta}{2}\right)
\end{pmatrix}
\end{equation}

Starting from the computational ground state $|0\rangle^{\otimes N}$, a rotation is applied independently to each qubit to encode the latent vector. The total encoding unitary can be expressed as:
\begin{equation}
U_{\text{enc}}(\mathbf{z}) = \bigotimes_{i=1}^{N} R_Y(\pi z_i)
\end{equation}
A scaling factor of $\pi$ is used to allow the latent variables to span the full range of rotation angles.

By applying the encoding unitary of the inital state, we prepare the quantum state $|\psi_{enc}\rangle$ , expressed as a tensor product of single qubit rotations.

\begin{equation}
|\psi_{\text{enc}}\rangle
= U_{\text{enc}}(\mathbf{z}) |0\rangle^{\otimes N}
= \bigotimes_{i=1}^{N}
\left(
\cos\left(\frac{\pi z_i}{2}\right)|0\rangle
+
\sin\left(\frac{\pi z_i}{2}\right)|1\rangle
\right)
\end{equation}

This expression explicitly represents the encoded state as a tensor product of single-qubit states. It highlights how each latent vector component controls the probability amplitudes of the corresponding qubit, thereby integrating continuous classical information into a quantum state. Angle encoding is particularly suitable for near-term quantum machine learning applications due to its low circuit depth and native hardware efficiency, while also enabling the encoding of continuous random variables without introducing additional trainable parameters. Furthermore, angle encoding maintains a structured representation within the learned latent space.

\subsubsection{Variational Layers with Successive Unitary Data Injection}

Variational layers form the trainable component of the quantum generator network that is responsible for converting encoded latent data into rich quantum states. While the encoding layer maps this latent vector into the quantum system using rigid data-dependent rotations, the use of variational layers adds trainable components in terms of adaptive rotations that depend on feedback from the discriminator network during adversarial learning.

Every layer in the variational part is made up of rotation gates on individual qubits and a multi-qubit entanglement operation. For an $N$-qubit quantum register, the unitary operator corresponding to a variational layer can be written as
\begin{equation}
U_{\mathrm{var}}(\boldsymbol{\theta})
=
U_{\mathrm{ent}}
\left(
\bigotimes_{i=1}^{N}
R_Z\!\left(\theta_i^{(z)}\right)
R_X\!\left(\theta_i^{(x)}\right)
\right),
\tag{5}
\end{equation}
where $R_X(\cdot)$ and $R_Z(\cdot)$ represent rotations about the Pauli-X and Pauli-Z axes, respectively.

$\boldsymbol{\theta} = \{ \theta_i^{(x)}, \theta_i^{(z)} \}_{i=1}^{N}$ indicates a set of parameters that require training. The fixed order of $R_X$ and $R_Z$ is maintained throughout and over all qubits and layers, providing enough degrees of freedom for each qubit to search over its state space while maintaining a hardware-efficient circuit arrangement.

The single-qubit rotation gates are given by
\begin{equation}\label{equation_Rx_theta_definition}
R_X(\theta) = e^{- i \frac{\theta}{2} X},
\quad
R_Z(\theta) = e^{- i \frac{\theta}{2} Z},
\tag{6}
\end{equation}
where $X$ and $Z$ denote the Pauli operators.

After the local rotations, an entanglement step is added to entangle the qubits. The entangling unitary transformation is accomplished through the use of CNOT gates in ring topology, and the entangling unitary transformation is given by
\begin{equation}
U_{\mathrm{ent}} =
\prod_{i=1}^{N}
\mathrm{CNOT}\!\left(i,\; (i \bmod N) + 1 \right).
\tag{7}
\end{equation}

In this way, it is possible to ensure that a communication-efficient connectivity pattern is provided, which will also be consistent with the conventional nearest-neighbor connectivity requirements of near-term quantum hardware.

The shortcoming of single-shot methods for data encoding in a quantum setup is that expressiveness emerges as a function of solely the available qubits, and hence such encodings are relatively less apt for handling large latent spaces. To circumvent the above constraint, in this proposed design of the quantum generator network, the idea of successive unitary data injection (SUDAI), wherein the process of data injection is staggered throughout different stages of the variational circuit, has been borrowed, inspired by previously established high-dimensional quantum GAN encoding \cite{Kalfon2024SuDaI}. In normal data injection, there is only a singular point of encoding of the latent vector in a circuit. However, in this proposed design, different stages of the circuit encode the same latent vector.

The overall generator unitary with $K$ successive injections can now be formally expressed as
\begin{equation}
U_G(\mathbf{z}, \boldsymbol{\theta})
=
\prod_{k=1}^{K}
\left[
U_{\mathrm{var}}^{(k)}(\boldsymbol{\theta}_k)\,
U_{\mathrm{enc}}(\mathbf{z})
\right],
\tag{8}
\end{equation}
where each variational layer $U_{\mathrm{var}}^{(k)}$ contains an independent set of parameters $\boldsymbol{\theta}_k$ that can be learned during training. This approach moves the expressiveness of the circuit from its width to depth, allowing a broader range of kets to be represented using a fixed number of qubits without disproportionately deepening each block of layers.

\begin{figure}[H]
    \centering
    \includegraphics[width=\linewidth]{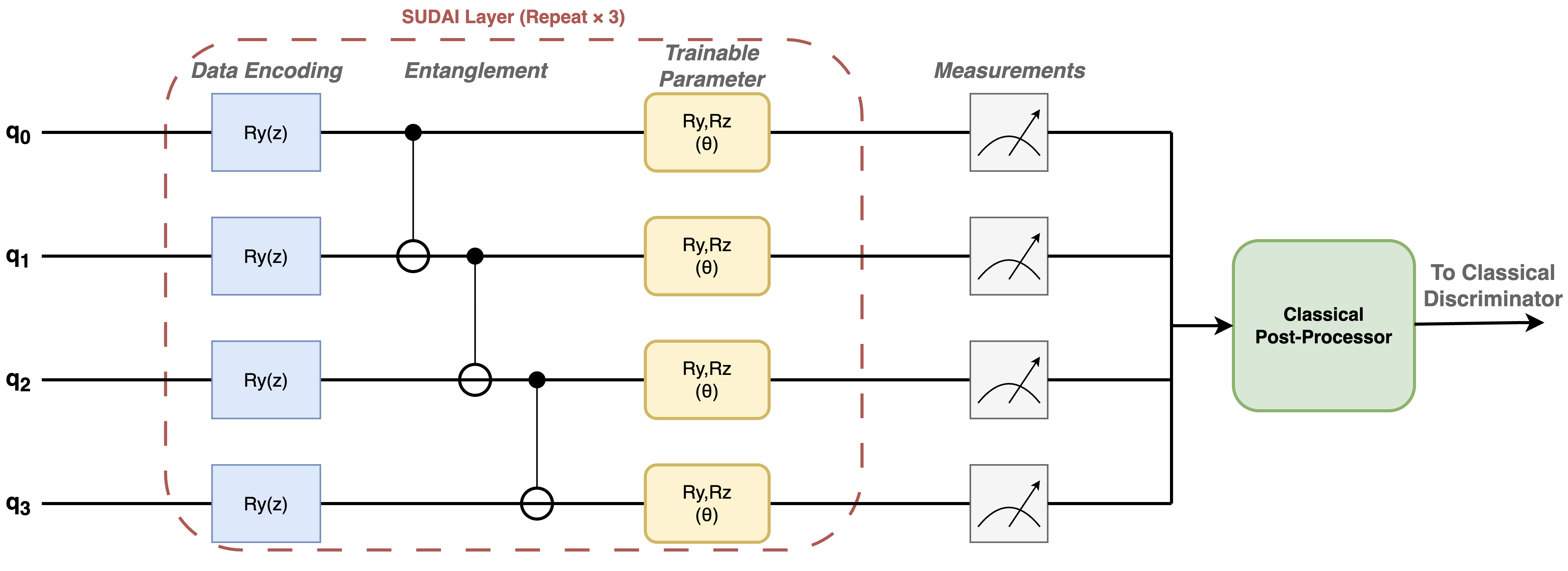}
    \caption{Quantum generator circuit using sequential unitary data injection (SUDAI). The latent noise vector is angle-encoded on four qubits via $R_Y(z_i)$ rotations, followed by nearest-neighbor entanglement and trainable variational rotations $R_Y(\theta)$ and $R_Z(\theta)$. The SUDAI block is repeated three times before measurement, and the resulting expectation values are forwarded to a classical post-processing network.}
    \label{fig:qgan_circuit}
\end{figure}

\subsubsection{Measurement Layer}

Finally, after the final SUDAI injection phase, the quantum circuit yields an evolved multi-qubit state that encodes the transformed latent information. To link this quantum representation to the classical components of the GAN, it is important for the quantum state to translate into a classical feature vector.

In our implementation, the measurement operation is carried out by calculating the expectation values for the Pauli $Z$ operator on each qubit. For the $N$-qubit system, the measurement process generates an $N$-dimensional real vector, the elements of which are the expectation values for individual qubits.

All expectation values are in the range $[-1, 1]$ and represent the weighted difference between the probabilities of a measurement of computational basis states $\lvert 0 \rangle$ and $\lvert 1 \rangle$. An expectation value close to $+1$ represents a bias toward $\lvert 0 \rangle$, while a value close to $-1$ represents a bias toward $\lvert 1 \rangle$. Values close to zero represent superpositions.

Altogether, the process of measurement results in a quantum feature vector
\begin{equation}
\mathbf{x} =
\left(
\langle Z_1 \rangle,
\langle Z_2 \rangle,
\ldots,
\langle Z_N \rangle
\right),
\tag{9}
\end{equation}
which represents the output generated by the quantum circuit. This vector contains information regarding the correlation and transformations acquired by the quantum circuit. It also maintains classical neural compatibility.

A benefit of using Pauli-$Z$ expectation values is that they form a stable readout scheme. It is also hardware-compatible, meaning it does not require full state tomography. The resulting quantum vector of features in the proposed QC-GAN architecture is followed by a classical post-processing network, where the measured values are refined before being assessed by the discriminator.

\subsubsection{Classical Discriminator Networks}
After measuring the quantum system, the expectation values from the generator become a classical feature vector. This feature vector is then used as input for the classical discriminator's task to differentiate between real data from the UNSW-NB15 set and the generated data from the quantum generator.

The discriminator is a simple fully connected feedforward network and operates strictly in the classical domain. The input dimension is the same as the number of qubits measured and provides an 8-dimensional feature vector. The network configuration includes two hidden layers consisting of 128 and 64 neurons respectively. The layers are each followed by batch normalization, Leaky ReLu activation functions, and dropout at a probability of 0.3.

The output neuron is a neuron with a sigmoid activation function that produces a value that lies in between 0 and 1 and represents how likely the input is from the original data distribution.  When training, the generator provides information on the quality of the generated patterns, allowing the gradual improvement of the generator through adversarial training. The choice of a classical discriminator saves additional quantum complexity and prevents the destabilization of gradients that could be problematic, especially when variational quantum circuits are more sensitive to noise.

\section{TRAINING AND RESULTS}

\subsection{Experimental Setup}

For comparison, we trained three different models: Classical GAN, QC-GAN without noise (clean), and QC-GAN with hardware noise simulation. All three models were trained on the UNSW-NB15 dataset, taking the four features selected in Section 2: synack, ct state ttl, sbytes, and smean. The training set contains 67,512 instances, the validation set has 14,820 instances, while the testing set holds 175,341 instances. Quantile transformation was used for each feature on training data only via QuantileTransformer.Our quantum generator uses 4 qubits, three SUDAI injection blocks, and two layers in each block -- 16 quantum parameters and a 356 parameter classical post-processor (372 total). The Classical GAN generator has 1,412 parameters, about 3.8 times more than the quantum generator. Discriminator for both models was spectral-normed and shared and had 8,961 parameters. We used the Wasserstein-GP loss function with $n_{critic} = 5$.Clean QC-GAN model used \texttt{lightning.qubit} from PennyLane with adjoint differentiation. For the noisy QC-GAN model, we used the \texttt{default.mixed} device with depolarizing, bit-flip, and amplitude-damping noise channels to simulate quantum hardware noise.

\subsection{Training Dynamics and Convergence}

\begin{figure}[htbp]
  \centering
  \includegraphics[width=\linewidth]{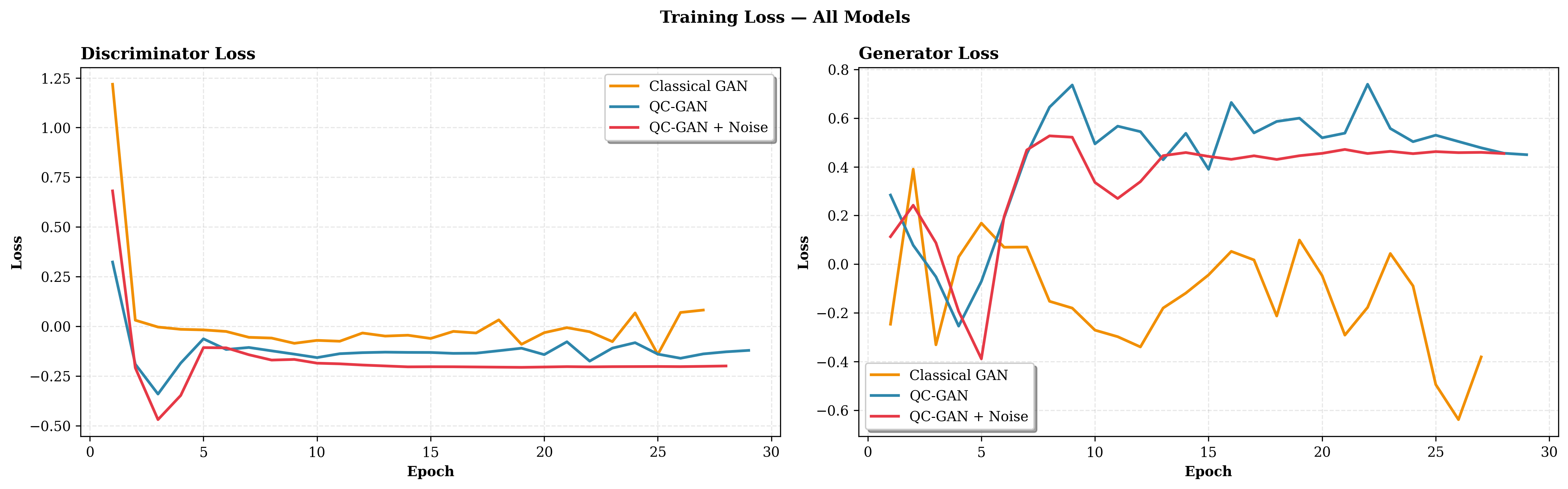}
  \caption{Here, we plot training loss for all three models. Left: Discriminator loss for QC-GAN variants stabilizes at
$-0.25$ after just a few epochs, while Classical GAN remains more noisy throughout training. Right: Generator loss for
both QC-GAN models converges to a constant positive value between $+0.4$ and $+0.5$; Classical GAN fluctuates more but
also stays positive. No signs of mode collapse in any model were observed.}
  \label{fig:loss}
\end{figure}

Figure~\ref{fig:loss} shows how the discriminator and generator losses change
across training for all three models.

As we can see, the discriminator loss for QC-GANs is dropping rapidly and then stays almost constant. This tells us
that our quantum generator reached stable equilibrium with the discriminator early in training. As expected, Classical GAN
discriminator is quite noisy and oscillates, which happens when a generator with many more parameters switches modes.
Generator loss for QC-GANs stabilizes in a certain range and stays there. On the contrary, Classical GAN generator
loses in the game with discriminator and its loss goes negative several times. In that case, it means that the generator wins
over the discriminator and does not generate instances that match training data distribution properly.

\begin{figure}[htbp]
  \centering
  \includegraphics[width=\linewidth]{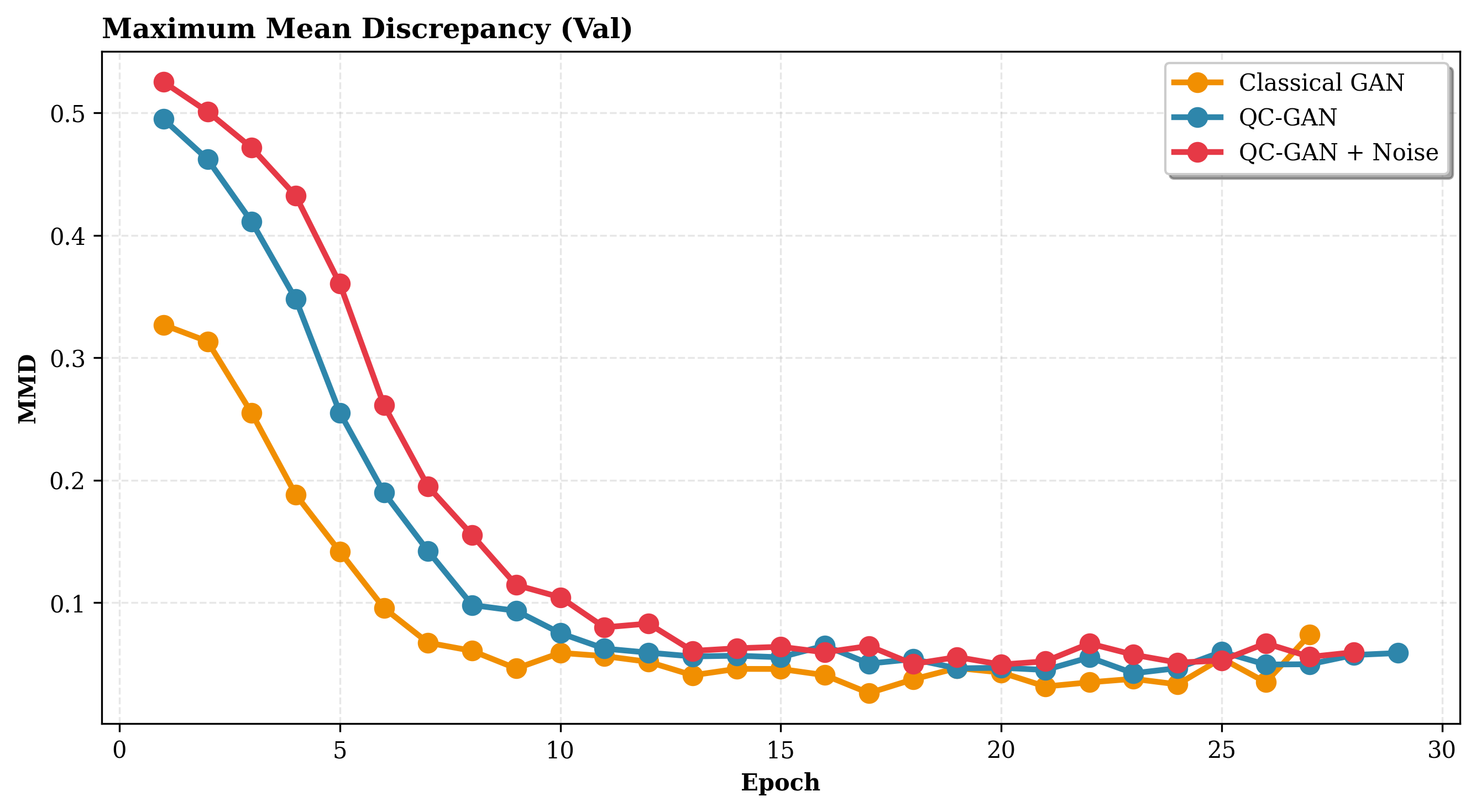}
  \caption{Validation MMD across training epochs.
           All three models start with high MMD and drive it down
           steadily.
           The Classical GAN converges fastest (best MMD $= 0.026$
           at epoch~17).
           The clean QC-GAN reaches best MMD $= 0.046$ at epoch~19,
           and QC-GAN\,+\,Noise reaches $0.050$ at epoch~18.
           All models learn to approximate the real traffic
           distribution, with the quantum models following a similar
           trajectory despite their much smaller parameter count.}
  \label{fig:mmd}
\end{figure}

Figure~\ref{fig:mmd} shows how well each generator matches the real data
distribution during training, measured by Maximum Mean Discrepancy (MMD).

However, all three models begin at MMD between 0.33 and 0.53 and then decrease consistently. The Classical
GAN converges fastest, which follows from its greater parameter budget. The clean QC-GAN model trails slightly,
and the noisy model converges slower in early epochs due to perturbation in gradient flow caused by noise
channels. Crucially, however, the latter does not get stuck or diverge. The convergence of the model demonstrates
that the SUDAI architecture is resilient against hardware-level noise.

\subsection{Generative Quality on the Test Set}

Table~\ref{tab:gen} reports the final generative quality metrics evaluated on
the held-out test set using the best checkpoint from validation.

\begin{table}[htbp]
\caption{Test-set generative quality metrics. WD and KL are mean $\pm$ std
         across test batches. Lower is better for all metrics.}
\label{tab:gen}
\centering
\begin{tabular}{lcccc}
\toprule
\textbf{Model (params)} & \textbf{MMD}$\downarrow$ & \textbf{MSE}$\downarrow$ &
\textbf{WD}$\downarrow$ & \textbf{KL}$\downarrow$ \\
\midrule
Classical GAN (1{,}412) & 0.3234 & 0.7508 & $0.112 \pm 0.045$ & $1.81 \pm 0.70$ \\
QC-GAN (372)            & 0.3619 & 0.7271 & $0.135 \pm 0.055$ & $3.01 \pm 2.99$ \\
QC-GAN\,+\,Noise (372)  & 0.3603 & \textbf{0.7033} & $0.140 \pm 0.062$ & $6.74 \pm 5.26$ \\
\bottomrule
\end{tabular}
\end{table}

The Classical GAN achieves the lowest MMD score (0.323). Consequently, it generates the distribution closest
to real traffic. More parameters mean higher representational capacity, and thus higher accuracy of distribution
approximation. However, looking at MMD does not provide a full picture because it measures similarity of
distributions in aggregate, and classifiers analyze individual samples, not entire distributions. The clean QC-GAN
gets a lower MSE score than the Classical GAN (0.727 vs. 0.751), and the noisy QC-GAN achieves the lowest
MSE (0.703). While MMD assesses distance between distributions as a whole, MSE compares generated samples
with real data points, meaning that it measures how well the generator creates individual flows that match
real traffic. Given the task of IDS evasion, MSE is therefore a more relevant metric, and in this respect, the
quantum generator outperforms the Classical GAN.

\begin{figure}[htbp]
  \centering
  \includegraphics[width=\linewidth]{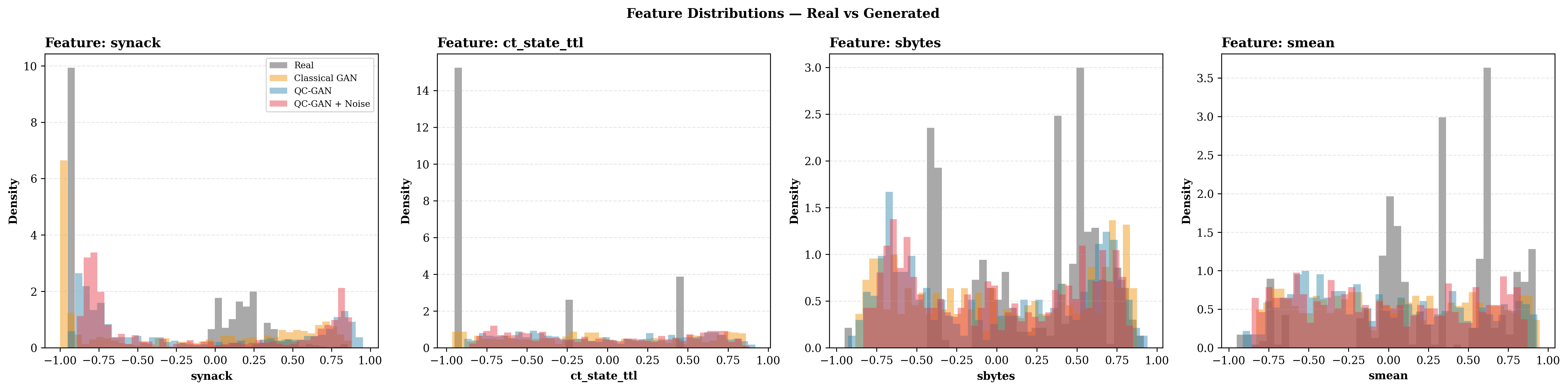}
  \caption{Per-feature distributions: real (grey) vs.\ Classical GAN
           (orange), QC-GAN (blue), and QC-GAN\,+\,Noise (red) across
           all four quantum features.
           All generators approximate the broad shape of the real
           distributions.
           The Classical GAN most closely matches the sharp peak near
           $-1.0$ in \texttt{synack} and \texttt{ct\_state\_ttl}.
           The QC-GAN\,+\,Noise model produces wider, more spread-out
           histograms across all features, consistent with noise
           acting as a regularizer that prevents the generator from
           collapsing to narrow regions.}
  \label{fig:dist}
\end{figure}

Figure~\ref{fig:dist} shows the per-feature distributions of generated
flows compared to real traffic across all four quantum features.

All three generators approximate the general shapes of distributions in all features. Classical GAN most accurately replicates the characteristic spike of real distribution at $-1.0$ for synack and ct state ttl. QC-GAN + Noise generates the noisiest distributions with the widest histograms, which corresponds with the idea of regularization through noise channels. For the two quantum features where the real distribution exhibits the sharpest peaks, all three models attempt to replicate this feature but fail to achieve the accuracy of the Classical GAN. On the other hand, in sbytes and smean, where real traffic forms a flat distribution, all models produce relatively accurate
approximations.

\subsection{IDS Evasion Results}

To test how successful the generated flows will be in tricking the IDS, we train classifiers Random Forest
(RF), XGBoost, and 1-D Convolutional Neural Network (CNN1D) on 16,000 balanced real samples (8,000
attack + 8,000 benign). Then each classifier was evaluated on 8,000 generated attack flows paired with 8,000
real benign flows.

The following metrics were used: Detection Rate (DR), Attack Success Rate ($\text{ASR} = 1 - \text{DR}$) and F1-score. Greater ASR means the IDS failed to detect a greater number of generated flows.

\begin{table}[htbp]
\caption{IDS evasion results. Higher ASR means more generated attack flows
        }
\label{tab:ids}
\centering
\begin{tabular}{llccc}
\toprule
\textbf{Model} & \textbf{IDS} & \textbf{DR}$\downarrow$ &
\textbf{ASR}$\uparrow$ & \textbf{F1}$\downarrow$ \\
\midrule
\multirow{3}{*}{Classical GAN}
  & RF      & 0.775 & 0.225 & 0.873 \\
  & XGBoost & 0.644 & 0.356 & 0.784 \\
  & CNN1D   & 0.712 & 0.288 & 0.832 \\
\midrule
\multirow{3}{*}{QC-GAN}
  & RF      & 0.820 & 0.180 & 0.901 \\
  & XGBoost & 0.589 & \textbf{0.411} & 0.741 \\
  & CNN1D   & 0.824 & 0.176 & 0.904 \\
\midrule
\multirow{3}{*}{QC-GAN\,+\,Noise}
  & RF      & 0.781 & 0.219 & 0.877 \\
  & XGBoost & 0.594 & \textbf{0.406} & 0.745 \\
  & CNN1D   & 0.769 & 0.231 & 0.869 \\
\bottomrule
\end{tabular}
\end{table}

\begin{figure}[htbp]
  \centering
  \includegraphics[width=\linewidth]{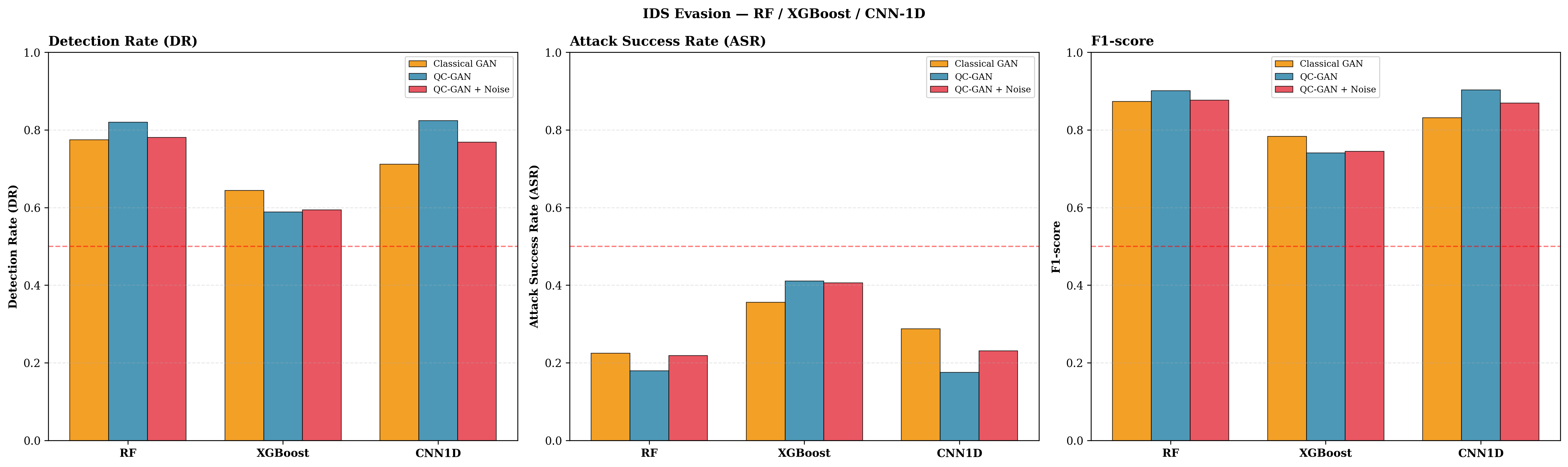}
  \caption{IDS evasion results across all models and classifiers.
           \textit{Left:} Detection Rate lower means fewer generated
           attacks were caught.
           \textit{Center:} Attack Success Rate higher means more
           generated flows evaded detection.
           \textit{Right:} F1-score of the IDS.
           Both quantum models outperform the Classical GAN against
           XGBoost, achieving ASRs of 41.1\% and 40.6\% vs.\ 35.6\%.}
  \label{fig:ids}
\end{figure}

The results in Table~\ref{tab:ids} and Figure~\ref{fig:ids} tell a clear
story.

In comparison to the Classical GAN, both quantum models outperform it on XGBoost. The
clean QC-GAN reaches ASR of 41.1\%, the noisy one - 40.6\%, while the Classical GAN scores 35.6\%. 
That means an improvement by 5.5 percentage points, despite 3.8$\times$ lower number of parameters.On CNN1D, the Classical GAN shows the highest ASR of 28.8\%. Meanwhile, the QC-GAN yields 17.6\%, and
the noisy variant -- 23.1%.
In contrast to the previous two models, against RF the difference between ASRs is negligibly low: 22.5\%
(Classical GAN), 18.0\% (clean QC-GAN), 21.9\% (noisy QC-GAN).

So, the quantum model does not always show superior performance. Yet, it beats other models where the task is the hardest, 
and this happens with orders of magnitude fewer parameters. 
This observation has interesting theoretical implications discussed later in this section.

\section{DISCUSSION}

\subsection{Why the Classical GAN Wins on Distribution Metrics and RF/CNN1D}

As we've seen, the Classical GAN achieves the best possible results measured by the largest MMD, smallest
Wasserstein distance and highest ASR in comparison to the rest of the models except for XGBoost, on which
the clean QC-GAN slightly edges out the Classical GAN. The Classical GAN simply has more parameters, and this increases its capability of imitating real attacks.
Classifiers such as RF or CNN1D are extremely sensitive to how well the underlying distribution of attack
flows is replicated.RF builds many independent decision trees based on random combinations of input features. Hence, it tends to
detect flows falling out of the distribution statistics. Because the Classical GAN manages to replicate the
distribution closely, its flows lie in the range of real attacks and escape detection.
Similar consideration applies to CNN1D. As a sequence-based model, it operates on feature values as a sequence,
detecting flows with similar pattern correlations between sequential elements. Better reproduction of the
distribution means better imitation of feature correlation, thus higher probability of evading detection.
Again, this doesn't mean that a quantum version of this model is somehow inferior to a Classical one.
Quite the opposite: when provided with four times more parameters, it becomes capable of learning
distribution in a way that is common practice for neural networks in classical machine learning tasks.

\subsection{Why the QC-GAN Wins on XGBoost}

XGBoost algorithm learns boosted decision trees which rely on threshold rules applied to individual or combined
features. As a consequence, it performs quite effectively in detecting flows landing within precise ranges of feature
values observed in real attack distributions.On account of accurate distribution replication, the Classical GAN produces flows lying in these dense regions
which are precisely the regions XGBoost knows how to classify into attack.The QC-GAN does not share this characteristic. Namely, the 4-qubit variational circuit works in a 16-
dimensional Hilbert space. Also, due to the SUDAI technique, it acquires enough degrees of freedom to
learn correlations of features in multiple passes through the encoder. Hence, the flows it generates become
statistically similar to real attacks (low MMD). Yet, they do not land within the most dense regions XGBoost
has detected because the quantum model discovers new parts of feature space not visited before.
\subsection{Why the Noisy QC-GAN Still Works}

Noise is normally considered an undesirable factor when applied to a quantum circuit. In this case, it turned out to be useful.
Channels of depolarizing, bit-flip, and amplitude-damping types push the generator toward more diffuse output modes. As a consequence, the model distributes its output flows over a wider range of feature space than those concentrated around attack zones having maximum density. Similar in function to dropout regularization in classical networks, noise helps avoid overfitting to narrowly defined modes. It is evident from the numbers.
The model with noise has the lowest test MSE (0.703) implying that its output flows are numerically closest to reality compared to other two models. In terms of XGBoost evasion, its performance reaches 40.6\%, which is only 0.5 percentage points less than in the case of the clean QC-GAN and exceeds the Classical GAN score by 5 points.
It should be remembered that in practice, an adversary employing a hybrid quantum-classical GAN does not need a perfect quantum computer but can rely on existing imperfect near-term hardware. The danger exists in the present, not some future time when fault-tolerant quantum computers emerge.

\subsection{Limitations}

There are a few limitations of this research to be mentioned. The evaluation involved only UNSW-NB15, a lab-generated dataset unlikely to represent completely the network traffic produced in production environments. The three IDS classifiers were standard off-the-shelf models without adversarial hardening -- otherwise the numbers of achieved ASRs could have been even lower. Lastly, the noise model is a simplification based on a channel approximation. It does not capture the entire range of noise factors in real-world quantum hardware including crosstalk and gate-specific errors.

\section{CONCLUSION}

In this paper, we demonstrated that it is possible to create an adversarial quantum-classical GAN that creates network flows evading classical IDSs using fewer parameters than in a classical GAN. The QC-GAN with only 372 parameters and a 4-qubit quantum circuit outperformed the Classical GAN (1,412 param-) on XGBoost, the strongest classifier among those included in this evaluation, by 5.5 percentage points in terms of attack success rate (41.1\% against 35.6\%). The higher ASR is associated with the ability of the quantum latent space to produce flows that fall outside decision boundaries of tree-based classifiers while being statistically realistic. As far as distribution-based evaluation is concerned, Classical GAN performs better because of a higher number of parameters. MMD and WD metrics confirm this advantage of Classical GAN, and also do RF and CNN1D classifiers.

The trade-off between a higher capacity for distribution representation and evasion capability is one of the major lessons learned in this research. The noisy QC-GAN indicates that imperfect near-term quantum hardware does not prevent successful evasion.
Under simulated noise conditions, the model exhibits both the lowest MSE and superior performance against XGBoost. Noise serves as a regularizer, spreading out the output distribution and making the flows more difficult to detect. In turn, the SUDAI circuit architecture turned out to be highly resistant to all training conditions considered. There was no mode collapse observed in any model, including the noisy one. These findings validate successive data injection as a valid approach to quantum generator architectures under near-term constraints.

All in all, these results indicate that the danger of a hybrid quantum-classical adversarial generator already exists at small scale of 4 qubits. In addition, the threat does not necessitate the use of fault-tolerant quantum computers. Future research will focus on testing the models on other datasets, examining different qubit counts, and IDSs hardened against GANs.
\section*{ACKNOWLEDGMENTS}

This work is partly sponsored by the National Science Foundation (NSF) 
awards numbers 2324924 and 2324925.
% References
\bibliographystyle{spiebib}
\bibliography{main}

@INPROCEEDINGS{11143115,
  author={Cirillo, Franco and Esposito, Christian},
  booktitle={IET Space and Communications Conference 2025}, 
  title={Intrusion detection using quantum generative adversarial networks: a federated approach with noisy simulators}, 
  year={2025},
  volume={2025},
  number={},
  pages={31-35},
  doi={10.1049/icp.2025.2226}
}

@article{VIBHUTE20242227,
title = {Network anomaly detection and performance evaluation of Convolutional Neural Networks on UNSW-NB15 dataset},
journal = {Procedia Computer Science},
volume = {235},
pages = {2227-2236},
year = {2024},
note = {International Conference on Machine Learning and Data Engineering (ICMLDE 2023)},
issn = {1877-0509},
doi = {https://doi.org/10.1016/j.procs.2024.04.211},
url = {https://www.sciencedirect.com/science/article/pii/S1877050924008871},
author = {Amol D. Vibhute and Minhaj Khan and Chandrashekhar H. Patil and Sandeep V. Gaikwad and Arjun V. Mane and Kanubhai K. Patel},
}

@misc{zolbayar2022generatingpracticaladversarialnetwork,
  title={Generating Practical Adversarial Network Traffic Flows Using NIDSGAN}, 
  author={Bolor-Erdene Zolbayar and Ryan Sheatsley and Patrick McDaniel and Michael J. Weisman and Sencun Zhu and Shitong Zhu and Srikanth Krishnamurthy},
  year={2022},
  eprint={2203.06694},
  archivePrefix={arXiv},
  primaryClass={cs.CR},
  url={https://arxiv.org/abs/2203.06694}
}

@article{Muneer2024,
  author = {Salman Muneer and Umer Farooq and Atifa Athar and Muhammad Ahsan Raza and Taher M. Ghazal and Shadman Sakib},
  title = {A Critical Review of Artificial Intelligence Based Approaches in Intrusion Detection: A Comprehensive Analysis},
  journal = {Journal of Engineering},
  volume = {2024},
  pages = {Article ID 3909173, 16 pp.},
  year = {2024},
  doi = {10.1155/2024/3909173},
  publisher = {Hindawi / John Wiley \& Sons}
}

@article{vieloszynski2024latentqganhybridqganclassical,
      title={LatentQGAN: A Hybrid QGAN with Classical Convolutional Autoencoder}, 
      author={Alexis Vieloszynski and Soumaya Cherkaoui and Ola Ahmad and Jean-Fr{\'e}d{\'e}ric Laprade and Oliver Nahman-L{\'e}vesque and Abdallah Aaraba and Shengrui Wang},
      year={2024},
      eprint={2409.14622},
      archivePrefix={arXiv},
      primaryClass={quant-ph},
      url={https://arxiv.org/abs/2409.14622}, 
}

@INPROCEEDINGS{10821224,
  author={Jha, Nitin and Parakh, Abhishek and Subramanian, Mahadevan},
  booktitle={2024 IEEE International Conference on Quantum Computing and Engineering (QCE)}, 
  title={A ML Based Approach to Quantum Augmented HTTP Protocol}, 
  year={2024},
  volume={02},
  number={},
  pages={591-592},
  doi={10.1109/QCE60285.2024.10420}
}

@article{Schuld2014IntroQML,
  author  = {Schuld, Maria and Sinayskiy, Ilya and Petruccione, Francesco},
  title   = {An Introduction to Quantum Machine Learning},
  journal = {Contemporary Physics},
  volume  = {56},
  number  = {2},
  pages   = {172--185},
  year    = {2014},
  month   = oct,
  doi     = {10.1080/00107514.2014.964942}
}

@article{Mirza2014CGAN,
  author       = {Mirza, Mehdi and Osindero, Simon},
  title        = {Conditional Generative Adversarial Nets},
  year         = {2014},
  journal      = {arXiv preprint arXiv:1411.1784},
  url          = {https://arxiv.org/abs/1411.1784}
}

@article{Huang2021QGAN,
  author    = {Huang, Kun and Wang, Zhi-An and Song, Cheng and Xu, Yirui and Deng, Hong and Rong, He-Liang and Su, Guang-Hui and Zheng, Hui and Zhang, Ying and Wang, Dong and Chen, Ming-Cheng and others},
  title     = {Quantum generative adversarial networks with multiple superconducting qubits},
  journal   = {npj Quantum Information},
  volume    = {7},
  number    = {1},
  pages     = {165},
  year      = {2021},
  month     = dec,
  publisher = {Nature Publishing Group},
  doi       = {10.1038/s41534-021-00503-1},
  url       = {https://doi.org/10.1038/s41534-021-00503-1}
}

@article{article1,
author = {Rath, Minati and Date, Hema},
year = {2024},
month = {10},
pages = {},
title = {Quantum data encoding: a comparative analysis of classical-to-quantum mapping techniques and their impact on machine learning accuracy},
volume = {11},
journal = {EPJ Quantum Technology},
doi = {10.1140/epjqt/s40507-024-00285-3}
}

@Article{math12213318,
AUTHOR = {Ranga, Deepak and Rana, Aryan and Prajapat, Sunil and Kumar, Pankaj and Kumar, Kranti and Vasilakos, Athanasios V.},
TITLE = {Quantum Machine Learning: Exploring the Role of Data Encoding Techniques, Challenges, and Future Directions},
JOURNAL = {Mathematics},
VOLUME = {12},
YEAR = {2024},
NUMBER = {21},
ARTICLE-NUMBER = {3318},
URL = {https://www.mdpi.com/2227-7390/12/21/3318},
ISSN = {2227-7390},
DOI = {10.3390/math12213318}
}

@article{Jha_2025,
   title={Toward a global quantum Internet: A review of challenges facing aerial quantum networks},
   volume={44},
   ISSN={1558-1772},
   url={http://dx.doi.org/10.1109/MPOT.2026.3657057},
   DOI={10.1109/mpot.2026.3657057},
   number={4},
   journal={IEEE Potentials},
   publisher={Institute of Electrical and Electronics Engineers (IEEE)},
   author={Jha, Nitin and Parakh, Abhishek},
   year={2025},
   month=jul,
   pages={6--12}
}

@misc{goh2025quantumenhancedgenerativeadversarialnetworks,
      title={Quantum-Enhanced Generative Adversarial Networks: Comparative Analysis of Classical and Hybrid Quantum-Classical Generative Adversarial Networks}, 
      author={Kun Ming Goh},
      year={2025},
      eprint={2508.09209},
      archivePrefix={arXiv},
      primaryClass={quant-ph},
      url={https://arxiv.org/abs/2508.09209}, 
}

@misc{nguemto2022reqganoptimizedadversarialquantum,
  title={Re-QGAN: an optimized adversarial quantum circuit learning framework}, 
  author={Sandra Nguemto and Vicente Leyton-Ortega},
  year={2022},
  eprint={2208.02165},
  archivePrefix={arXiv},
  primaryClass={quant-ph},
  url={https://arxiv.org/abs/2208.02165}
}

@misc{rao2023learningharddistributionsquantumenhanced,
  title={Learning hard distributions with quantum-enhanced Variational Autoencoders}, 
  author={Anantha Rao and Dhiraj Madan and Anupama Ray and Dhinakaran Vinayagamurthy and M. S. Santhanam},
  year={2023},
  eprint={2305.01592},
  archivePrefix={arXiv},
  primaryClass={quant-ph},
  url={https://arxiv.org/abs/2305.01592}
}

@article{Shu2024VQCGAN,
  title = {Variational Quantum Circuits Enhanced Generative Adversarial Network},
  author = {Shu, Runqiu and Xu, Xusheng and Yung, Man-Hong and Cui, Wei},
  journal = {arXiv preprint arXiv:2402.01791},
  year = {2024},
  url = {https://arxiv.org/abs/2402.01791}
}

@misc{he2025qganbaseddataaugmentationhybrid,
  title={QGAN-based data augmentation for hybrid quantum-classical neural networks}, 
  author={Run-Ze He and Jun-Jian Su and Su-Juan Qin and Zheng-Ping Jin and Fei Gao},
  year={2025},
  eprint={2505.24780},
  archivePrefix={arXiv},
  primaryClass={cs.LG},
  url={https://arxiv.org/abs/2505.24780}
}

@inproceedings{Stein_2021,
  title={QuGAN: A Quantum State Fidelity-based Generative Adversarial Network},
  url={http://dx.doi.org/10.1109/QCE52317.2021.00023},
  DOI={10.1109/qce52317.2021.00023},
  booktitle={2021 IEEE International Conference on Quantum Computing and Engineering (QCE)},
  publisher={IEEE},
  author={Stein, Samuel A. and Baheri, Betis and Chen, Daniel and Mao, Ying and Guan, Qiang and Li, Ang and Fang, Bo and Xu, Shuai},
  year={2021},
  month=oct,
  pages={71--81}
}

@article{Hammami2025QGANAnomaly,
  author       = {Hammami, Wajdi and Cherkaoui, Soumaya and Wang, Shengrui},
  title        = {Enhancing Network Anomaly Detection with Quantum GANs and Successive Data Injection for Multivariate Time Series},
  journal      = {arXiv preprint arXiv:2505.11631},
  year         = {2025},
  url          = {https://arxiv.org/abs/2505.11631},
  doi          = {10.48550/arXiv.2505.11631}
}

@article{hammami2025quantumgrugan,
  author       = {Hammami, Wajdi and Cherkaoui, Soumaya and Laprade, Jean-Frederic and Ahmad, Ola and Wang, Shengrui},
  title        = {Quantum Gated Recurrent GAN with Gaussian Uncertainty for Network Anomaly Detection},
  journal      = {arXiv preprint arXiv:2510.26487},
  year         = {2025},
  url          = {https://arxiv.org/abs/2510.26487},
  doi          = {10.48550/arXiv.2510.26487}
}

@article{rahman2024quantumintrusion,
  title={Quantum Generative Adversarial Networks for Intrusion Detection: Leveraging Quantum State Encoding for Complex Attack Patterns},
  author={Rahman, Md and Alshareef, Hani and Kim, Youseok},
  journal={IEEE Access},
  volume={12},
  pages={1--12},
  year={2024},
  publisher={IEEE},
  doi={10.1109/ACCESS.2024.XXXXX}
}

@article{Jha2024,
  title        = {Joint encryption and error correction for secure quantum communication},
  author       = {Jha, Nitin and Parakh, Abhishek and Subramaniam, Mahadevan},
  journal      = {Scientific Reports},
  volume       = {14},
  number       = {1},
  pages        = {24542},
  month        = {Oct},
  year         = {2024},
  issn         = {2045-2322},
  doi          = {10.1038/s41598-024-75212-8},
  url          = {https://doi.org/10.1038/s41598-024-75212-8}
}

@inproceedings{Moustafa2015UNSWNB15,
  author    = {Moustafa, Nour and Slay, Jill},
  title     = {UNSW-NB15: A Comprehensive Data Set for Network Intrusion Detection Systems (UNSW-NB15 Network Data Set)},
  booktitle = {Military Communications and Information Systems Conference (MilCIS)},
  year      = {2015},
  publisher = {IEEE},
  pages     = {}
}

@article{Moustafa2016EvaluationUNSW,
  author  = {Moustafa, Nour and Slay, Jill},
  title   = {The Evaluation of Network Anomaly Detection Systems: Statistical Analysis of the UNSW-NB15 Dataset and the Comparison with the KDD99 Dataset},
  journal = {Information Security Journal: A Global Perspective},
  year    = {2016},
  pages   = {1--14}
}

@article{Moustafa2017GeometricArea,
  author  = {Moustafa, Nour and others},
  title   = {Novel Geometric Area Analysis Technique for Anomaly Detection Using Trapezoidal Area Estimation on Large-Scale Networks},
  journal = {IEEE Transactions on Big Data},
  year    = {2017}
}

@incollection{Moustafa2017BigData,
  author    = {Moustafa, Nour and others},
  title     = {Big Data Analytics for Intrusion Detection System: Statistical Decision-Making Using Finite Dirichlet Mixture Models},
  booktitle = {Data Analytics and Decision Support for Cybersecurity},
  publisher = {Springer, Cham},
  year      = {2017},
  pages     = {127--156}
}

@inproceedings{Sarhan2020NetFlow,
  author    = {Sarhan, Mohanad and Layeghy, Siamak and Moustafa, Nour and Portmann, Marius},
  title     = {NetFlow Datasets for Machine Learning-Based Network Intrusion Detection Systems},
  booktitle = {Big Data Technologies and Applications: 10th EAI International Conference, BDTA 2020, and 13th EAI International Conference on Wireless Internet, WiCON 2020, Virtual Event, December 11, 2020, Proceedings},
  publisher = {Springer Nature},
  year      = {2020},
  pages     = {117}
}

@article{Kalfon2024SuDaI,
  author  = {Kalfon, Benjamin and Cherkaoui, Soumaya and Laprade, Jean-Fr{\'e}d{\'e}ric and Ahmad, Ola and Wang, Shengrui},
  title   = {Successive Data Injection in Conditional Quantum {GAN} Applied to Time Series Anomaly Detection},
  journal = {IET Quantum Communication},
  year    = {2024},
  volume  = {},
  number  = {},
  pages   = {},
  doi     = {10.1049/qtc2.12088}
}

\end{document}